\documentclass[runningheads]{llncs}

\usepackage{graphicx}
\usepackage{amsmath,amssymb} 
\usepackage{booktabs}        
\usepackage{hyperref}        
\usepackage[misc]{ifsym}     
\usepackage{xcolor}          
\usepackage{float}

\begin{document}

\title{Leveraging Vision-Language Models to Detect Attention in Educational Videos}

\titlerunning{Leveraging Vision-Language Models to Detect Attention}

\author{
 Gabriel Becquet\inst{1} \and
 S\'ebastien Lall\'e\inst{1} \and
 Vanda Luengo\inst{1} \and
 Ali Abou-Hassan\inst{1,2}}
\authorrunning{Becquet et al.}

\institute{
\textsuperscript{1}Sorbonne University, CNRS, LIP6 \& PHENIX, F-75005 Paris, France\\
\textsuperscript{2}Institut Universitaire de France (IUF), 75231 Paris, France\\
 \email{\{gabriel.becquet, sebastien.lalle, vanda.luengo, ali.abou\_hassan\}@sorbonne-universite.fr}
}

\maketitle

\begin{abstract}
Educational videos are a cornerstone of remote and blended learning. However, learners' fluctuating attention remains a significant barrier to effective information retention. Prior research has attempted to mitigate this by detecting and reacting to attention loss at runtime using eye tracking. Such detection has been based so far on classical machine learning classifiers trained on engineered features, such as summary statistics over learners' fixations and saccades. These methods have struggled to capture the complex, temporal nature of learner engagement, thus exhibiting moderate prediction performance.
In this study, we aim to advance the detection of attention by shifting from standard engineered features to a multimodal foundation models. Using an educational eye-tracking dataset (N = 70), we investigate a novel methodology that utilizes a Vision-Language Model (VLM) to analyze video content directly with superimposed gaze data. This approach aims to leverage the semantic reasoning capabilities of foundation models to contextualize learner focus within the video stream. We evaluate the performance of this VLM-based approach using several prompting strategies with Gemini 3, but ultimately found that none of them could outperform statistical baselines. Our results provide new insights into the limitations of using VLMs for real-time educational diagnostics.

\end{abstract}

\section{Introduction}

In the landscape of remote and blended education, instructional videos have established themselves as a fundamental delivery mechanism \cite{lalle2025}. While effective when best practices are followed, video-based learning often suffers from a critical limitation: the fluctuation of learner attention. Research indicates that attention can vary significantly throughout a video due to both internal states (e.g., fatigue) and external factors (e.g., content difficulty) \cite{faber2020,kuvar2023detecting}. Detecting these fluctuations in real-time is crucial, as it opens the door to adaptive systems that can intervene to re-engage the learner or reinforce key concepts.

To address this challenge, the field of Artificial Intelligence in Education (AIED) has increasingly turned to eye-tracking technology \cite{bixler2021,hutt2017,kuvar2023detecting,lalle2025}. The prevailing approach in the current literature relies on a pipeline of "classical" machine learning (ML): extracting engineered features from raw gaze data, such as fixation duration, saccade velocity, and pupil dilation, and feeding them into classifiers like Random Forests or Support Vector Machines. While recent advancements have attempted to enrich these models by incorporating dynamic Areas of Interest (AOIs) to capture what specifically the student is looking at (e.g., a formula vs. a decorative image) \cite{lalle2025}, these methods face inherent scalability issues. Defining AOIs often requires labor-intensive manual annotation or complex object detection pipelines that may not generalize across different video domains. Furthermore, engineered features often reduce complex, temporal viewing behaviors into static summary statistics, potentially discarding valuable semantic context about the interaction between the learner's gaze and the visual content. Another issue is that the existing datasets leveraged in related work tend to be rather small in magnitude, due to the cost of collecting large eye-tracking datasets, which makes it very difficult to train more advanced ML models such as recurrent neural networks. 

In this paper, we propose a paradigm shift from manual feature engineering to end-to-end multimodal understanding. We introduce a novel methodology utilizing Vision-Language Models (VLMs) to detect attention. Unlike classical approaches that require explicit definitions of \textit{where} a student should look, VLMs can leverage their pre-trained semantic knowledge to reason about the visual scene directly. By superimposing gaze data onto video frames, we enable the model to "see" the learner's focus in context, effectively asking the model to interpret whether the gaze location aligns with the pedagogical intent of the moment. Another major advantage of VLMs is their ability to bypass the collection of extensive eye-tracking training data, which is a costly and difficult process to implement in educational contexts as discussed earlier.

Using an eye-tracking dataset collected from $N=70$ students in a chemistry course, we evaluate the efficacy of this VLM-based approach against statistical baselines. Our contributions are twofold:
\begin{enumerate}
    \item We present a technical framework for transforming eye-tracking data into visual prompts suitable for VLMs, bypassing the need for manual AOI definition as well as task-specific training data collection.
    \item We provide an empirical comparison between different VLMs' prompting strategies and baselines, offering insights into the feasibility of using foundation models for real-time educational diagnostics.
\end{enumerate}

\section{Related Work}

\textbf{Eye-Tracking for Attention and Mind-Wandering Detection.}
A large body of work links eye movements to attention and mind-wandering.
In reading, gaze behavior robustly correlates with comprehension and mind wandering \cite{dmello2016,faber2020}.
In video lecture contexts, detection is typically harder and performance is more modest \cite{kuvar2023detecting}.
For example, Hutt et al.\ trained student-independent models for lecture viewing and reported that global gaze features outperformed stimulus-dependent local (AOI/grid-based) features (F1$_{\text{MW}}$ = 0.47 vs.\ 0.36; chance F1$_{\text{MW}}$ = 0.30) \cite{hutt2017}.
Zhao et al.\ proposed scalable MOOC mind-wandering detection using webcam-based gaze estimation and compared global vs.\ AOI-based local features; their results show that performance depends on the feature set and setup, with best reported F1 around 0.41 on their data \cite{yue2017scalable}.
More recently, Lall\'e et al.\ introduced dynamic AOIs and feature fusion for educational videos, reporting improved student-independent detection of attention-related states in this setting \cite{lalle2025}.
Across this literature, most approaches rely on coarse or manually defined AOIs and handcrafted statistical features paired with conventional classifiers, and generalization across domains remains challenging \cite{bixler2021}.

\textbf{Vision-Language Models (VLMs) in Education.}
Deep learning has transformed computer vision, but educational eye-tracking datasets remain too small for these approaches, e.g., a few hundred data points in \cite{bixler2021,hutt2017,lalle2025}.
In parallel, modern multimodal foundation models can jointly process visual context and language \cite{wu2023large}, which motivates exploring if they can interpret learner attention more semantically (e.g., relating gaze position to what is being explained).

\textbf{Foundation Models as (Training-Free) Classifiers via Visual Prompting.}
Recent work shows that large pretrained multimodal models can often be steered without weight updates, through prompting and input editing \cite{shtedritski2023}.
Surveys of video-LLMs summarize this emerging paradigm for video understanding tasks more broadly \cite{wu2023large}.
Within gaze+VLM research, Madinei et al.\ propose a training-free use of real-time eye-tracking to disambiguate referents for VQA in multimodal LLMs \cite{madinei2025gaze}, while Mathew et al.\ introduce a VLM tailored for multi-task gaze understanding \cite{mathew2025gazevlm}.
Inspired by this line of work, our study evaluates whether \emph{Gemini 3} can diagnose learner attention from lecture video frames when gaze is provided as a visual prompt, without task-specific fine-tuning \cite{gemini3_blog,gemini3_api}.

\section{Methodology}

We frame the problem as a binary classification task: predicting \textit{Low vs. High Attention} based on a short window of eye-tracking data.

\subsection{Dataset and Preprocessing}
The dataset we used was previously collected by Lallé et al.~\cite{lalle2025} and shared publicly\footnote{\url{https://gitlab.lip6.fr/mocah-public/eye_tracking_green_chem_video}}. The data collection protocol is fully described in their paper, and here we provide an overview sufficient for the purpose of this experiment.

The dataset includes eye-tracking data from $N=70$ undergraduate students watching a 7-minute introductory green chemistry course. After signing a consent form and undergoing calibration of a Tobii Nano eye-tracker, participants watched the video alone in a room with constant lighting, without restrictions (e.g., they could skip, take notes). Next, participants completed an attention survey asking them to rate their level of attention at specific time points in the video, as well as a factual knowledge recall test, composed of questions assessing recall of factual information from the video\footnote{Not used in this study but administered to triangulate and validate participants’ self-reported attention ratings in the attention survey.}. Participants’ responses to the attention survey are used to derive the labels for the aforementioned binary classification task. All collected data were kept fully anonymous. The protocol was assessed by the university's data protection officer and ethics board.

To construct the binary classification target, we thresholded the self-reports: ratings of 0--2 were labelled as "Inattentive" (Class 0), and ratings of 3--5 were labelled as "Attentive" (Class 1). This resulted in a class distribution of 19.9\% Inattentive and 80.1\% Attentive.
We synchronized the different data channels by mapping the gaze data and the labels to specific video frames (corresponding to the time points targeted in the survey items) based on UTC timestamps in microseconds.

\subsection{Multimodal Reasoning with VLMs}
We propose a novel method using the Gemini 3 VLM to classify learner state by processing the entire video segment. To reduce inference cost and latency when processing many video clips, we used the Gemini 3 Flash variant, which is optimized for high-throughput multimodal inference.

\begin{figure}[t]
  \centering
  \includegraphics[width=0.8\textwidth]{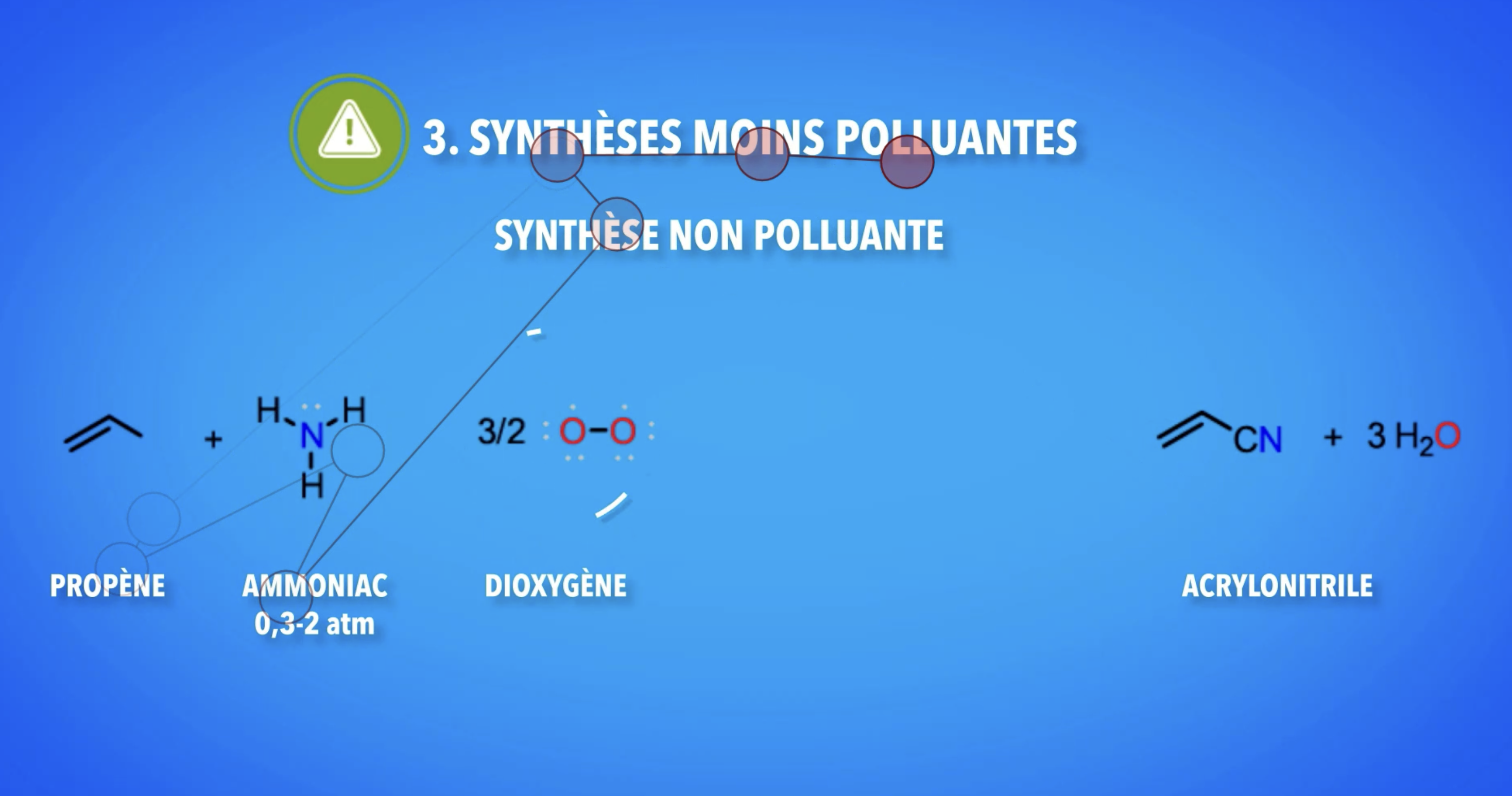}
  \caption{Example of visual prompting: The student's gaze red circle is superimposed onto the educational video frame. This allows the VLM to spatially correlate the gaze location with specific visual content.}
  \label{fig:gaze_overlay}
\end{figure}

\textbf{Video Generation (Visual Prompting).}
Following \cite{lalle2025}, we focused on 20-second video segments that correspond to the 20s before the specific time points of the self-reports survey. For each segment, we generated a corresponding video clip. We superimposed the student's gaze trajectory onto the video frames as a dynamic, semi-transparent red circle (see Fig.~\ref{fig:gaze_overlay}). This ensured the model perceives both the visual content (the lecture slides, texts, images, diagrams, tables) and the temporal gaze behaviors (saccades, fixations, and scanning patterns) simultaneously.


\textbf{Prompting Strategies.}
We evaluated several prompting strategies via the Gemini 3 API\footnote{The actual content of the prompts is publicly shared: \url{https://osf.io/8ywn3/files/hjgdy?view_only=dd0406a9a2d9427ea6c6f5bd86dee66d}.}:
\begin{itemize}
    \item \emph{Direct Classification:} The model is provided with the video and asked to output "Attentive" or "Inattentive" directly. This initial approach was tested but excluded in this study due to poor performance.
    \item \emph{Heuristic Prompting (Chain-of-Thought):} We instructed the model to follow a reasoning pipeline: (1) Identify the key pedagogical concept on screen; (2) Estimate a "Gaze-Content Alignment" score (0-100); (3) Classify the state based on this alignment. This strategy aims to force the model to explicitly reason about the gaze location before committing to a label.
    \item \emph{Few-Shot:} We provided video-based examples of "Attentive" and "Inattentive" scenarios to guide the model (either 1 or 5 examples per class). 
    \item \emph{Blind similarity:} We provided video-based examples of "Attentive" and "Inattentive" scenarios to guide the model, but we label them as class A and B instead of saying what is attentive or not. This is to explore whether attention classification can be better achieved through visual similarity rather than relying on explicit semantic labels.
\end{itemize}

\section{Results}

We evaluated the performance of the Gemini 3 VLM across the distinct prompting strategies defined above on a dataset of $N=1,033$ video segments. Recall that the dataset exhibits a significant class imbalance, with 80.1\% of segments labelled as "Attentive" and 19.9\% as "Inattentive" (Class 0). Table~\ref{tab:main_results} compares our VLM approaches against three statistical baselines: Uniform Random, Proportional Random, and Majority Class.

\subsection{Performance vs. Baselines}
The results indicate that off-the-shelf Visual Prompting with Gemini 3 faces challenges in this domain compared to the \textit{Majority Class} baseline, which achieves 80.1\% accuracy simply by predicting "Attentive" for every student (hence a lower F1-score of 0.45). The VLM strategies achieved accuracies between 58\% and 60\% (F1 from 0.48 to 0.50), struggling to overcome the strong class imbalance without frequent false positives.

However, a qualitative distinction is observed in the \textbf{Macro metrics}. The \textit{Zero-shot (Heuristics)} strategy achieved the highest \textbf{Macro Recall} (0.525) and \textbf{Macro Precision} (0.517) across all methods. While these numerical improvements over the random baselines (0.50) are marginal given the sample size ($N \approx 1000$), they signal a fundamental difference in classifier behavior. 

First, the precision of 0.517 is above the Proportional Random baseline (0.50), suggesting that the model's Chain-of-Thought reasoning provides a slight discriminative edge over pure chance. Second, regarding recall, the Majority baseline achieves its score by ignoring the "Inattentive" class entirely (Class 0 Recall = 0.0). In contrast, the VLM actively identifies instances of inattention. For an educational diagnostic system, a classifier that attempts to detect struggling students, with precision better than random guessing, provides a starting point for intervention that a "blind" majority classifier cannot offer. The other baselines do not suffer from this issue, but generate slightly more false negatives than the \textit{Zero-shot (Heuristics)} strategy.

\begin{table}[t!]
\centering
\caption{Classification performance comparison ($N \approx 1000$ video segments). The \textit{Zero-shot (Heuristic)} strategy achieves the highest Macro Recall (0.525) and Macro Precision (0.517) among VLM methods, surpassing the baselines in detecting the minority class.}
\label{tab:main_results}
\setlength{\tabcolsep}{4pt} 
\begin{tabular}{lcccc}
\toprule
\textbf{Method} & \textbf{Acc.} & \textbf{Prec. (Mac)} & \textbf{Rec. (Mac)} & \textbf{F1 (Mac)} \\
\midrule
\textit{Baselines} & & & & \\
Majority Class (All "Attentive") & \textbf{0.801} & 0.400 & 0.500 & 0.445 \\
Proportional Random ($P=Freq$) & 0.681 & 0.501 & 0.501 & \textbf{0.500} \\
Uniform Random ($P=0.5$) & 0.500 & 0.500 & 0.500 & 0.450 \\
\midrule
\textit{VLM Strategies} & & & & \\
Similarity Matching (Blind) & 0.597 & 0.494 & 0.492 & 0.483 \\
Few-shot (1 Exemplar) & 0.606 & 0.503 & 0.504 & 0.493 \\
Few-shot (5 Exemplars) & 0.604 & 0.504 & 0.506 & 0.494 \\
Zero-shot (Heuristic CoT) & 0.594 & \textbf{0.517} & \textbf{0.525} & \textbf{0.500} \\
\bottomrule
\end{tabular}
\end{table}

\subsection{Prompting Strategy Analysis}
Comparing the VLM strategies reveals that \textbf{Few-shot prompting did not yield significant improvements} over Zero-shot heuristics. While the \textit{One-shot} approach was marginally the best performing VLM method in terms of raw accuracy (60.6\%), adding more exemplars (5-shot) caused performance to plateau. 
This counter-intuitive result implies that the model's difficulty lies not in understanding the task definition (which exemplars typically resolve), but in the inherent ambiguity of interpreting gaze circles without broader temporal or pedagogical context. The superior Macro Recall of the \textit{Zero-shot Heuristic} method further suggests that forcing the model to explicitly reason about "Gaze-Content Alignment" via a Chain-of-Thought process is more robust than simple similarity matching against few-shot examples. This reinforces our finding that the value of VLMs in this domain lies in their semantic reasoning capabilities (interpretability) rather than in pattern matching raw visual signals.

\subsection{The Value of Interpretability}
A critical advantage of the VLM approach, which is not captured by standard quantitative metrics, is \textbf{interpretability}. Unlike classical classifiers (e.g., SVMs, Random Forests) that output a single probability score, and possibly feature importance scores, based on engineered feature vectors, our VLM pipeline generates a natural language justification for every prediction.

For instance, in a datapoint classified as \textbf{"Attentive"}, the model reasoned:
\begin{quote}
\textit{"The student shows clear semantic alignment by focusing on the hazard symbols of the product DDT. This behavior suggests they are actively attending to the safety and environmental implications of the chemical synthesis being presented on the static slide."}
\end{quote}
The model further cited specific visual evidence to support this claim: \textit{"The gaze cursor moves from the 'Reagent' label towards the hazard symbols... [and] remains focused on the hazard symbols for approximately 15 seconds."}

Conversely, for a datapoint classified as \textbf{"Inattentive"}, the model noted:
\begin{quote}
\textit{"The student's gaze behavior is highly erratic and non-systematic, frequently moving to the periphery of the screen rather than following the instructional flow. This pattern is strongly similar to the inattentive exemplar and indicates a lack of engagement with the material."}
\end{quote}

Evidence cited included: \textit{"Gaze cursor frequently jumps to screen margins... Movement across text and formulas is erratic... Lack of sustained focus on key visual elements."}

In these explanations, the VLM appears to be successful in mapping students' gaze to the visual stimuli in the video, suggesting that its lack of accuracy was more likely due to the difficulty of linking these gaze patterns to internal disengagement states. Of course, more formal investigations of these justifications are required (e.g., with human annotators) to actually gauge their validity.
Nonetheless, this granular, semantic reasoning could allow educators and system designers to understand \textit{why} a student was flagged, offering actionable insights for intervention design that features based on summary statistics (e.g., fixation rate, saccade velocity) cannot readily provide.

\section{Conclusion and Future Work}

This study investigated the feasibility of using a general-purpose Vision-Language Model (Gemini 3) as a training-free classifier for learner attention. By superimposing gaze circles onto educational videos, we tested whether visual prompting alone could bridge the "semantic gap" in eye-tracking analysis.

Our evaluation ($N \approx 1000$) reveals that while the approach is technically viable, off-the-shelf VLMs (more specifically Gemini 3) currently struggle to outperform even statistical baselines in raw accuracy. The \textit{Zero-Shot Heuristic} strategy achieved the highest \textbf{Macro Recall (0.525) and Precision (0.517)}, but these performances are still substantially lower than traditional ML classifiers that achieved an F1 of 0.59 in previous work on the same dataset~\cite{lalle2025}.
All prompting approaches overall exhibited a bias toward the majority "Attentive" class, suggesting that zero-shot and few-shots reasoning is insufficient to fully map subtle gaze trajectories to internal cognitive states without task-specific adaptation. Although this highlights a limitation of VLMs and their current inadequacy for the task, we view this finding as valuable to better document what VLMs can and cannot do in educational applications. Indeed, while LLMs and VLMs are increasingly used in education, training-free foundation models cannot always replace domain-specific modeling, which we found is very much the case for video-based attention detection. We also provide a concrete methodology for using eye-tracking data that did not perform well in this context, indicating that future research should pursue alternative ways of integrating foundation models for attention detection.

However, a promising finding lies in the fact that the VLM provided \textbf{natural language justifications} for its predictions. Unlike numerical classifiers, the model could articulate \textit{why} a gaze pattern was problematic (e.g., "fixating on a decorative element while a formula is explained"), offering a new layer of interpretability for educational diagnostics. Albeit investigating the quality and validity of these justifications is still future work, this interpretability capability may increase user trust and support teacher decision-making and model refinement when combined with alternative approaches that achieve stronger raw performance. This finding also suggests that while Gemini 3 could not accurately diagnose lapses in attention, it might be more suited to address other tasks such as analyzing visual exploration strategies.

We conclude that while Gemini 3 cannot yet replace specialized classifiers zero-shot in our context, it can offer novel semantic insights. Future work should move beyond prompting strategies, for instance toward Parameter-Efficient Fine-Tuning (PEFT) to adapt VLMs to the specific temporal dynamics of gaze, or Hybrid Architectures that combine the signal precision of feature engineering with the semantic reasoning of LLMs/VLMs.

To facilitate future research, we have released the specific VLM prompts used for all strategies, and an anonymized sample video stimuli on the Open Science Framework (OSF). These resources are available at: \url{https://osf.io/8ywn3/overview?view_only=dd0406a9a2d9427ea6c6f5bd86dee66d}.\\

\textbf{Acknowledgements.} This work was supported by the French National Research Agency (Grants ANR-22-CE33-0016 \&  ANR-23-IACL-0007). 

\bibliographystyle{splncs04}
\bibliography{references}

\end{document}